\documentclass[sigconf,review=false]{acmart}
\renewcommand\footnotetextcopyrightpermission[1]{} 
\usepackage{booktabs} 
\usepackage[utf8]{inputenc}
\usepackage{algorithm}
\usepackage{algpseudocode}
\usepackage[utf8]{inputenc} 
\usepackage{graphicx} 
	\usepackage{multirow}
\usepackage{romannum}

\newcommand{\etal}{et. al.}

\setcopyright{none}

\begin{document}
\title{RULLS: Randomized Union of Locally Linear Subspaces for Feature Engineering}

\author{Namita Lokare}
\affiliation{%
  \institution{SAS Institute}
  \streetaddress{100 SAS Campus Drive}
  \city{Cary}
  \state{North Carolina}
  \postcode{27513}
}
\email{namita.lokare@sas.com}

\author{Jorge Silva}
\affiliation{%
  \institution{SAS Institute}
  \streetaddress{100 SAS Campus Drive}
  \city{Cary}
  \state{North Carolina}
  \postcode{27513}
}
\email{jorge.silva@sas.com}

\author{Ilknur Kaynar Kabul}
\affiliation{%
  \institution{SAS Institute}
  \streetaddress{100 SAS Campus Drive}
  \city{Cary}
  \state{North Carolina}
  \postcode{27513}
}
\email{ilknur.kaynarkabul@sas.com}

\renewcommand{\shortauthors}{N. Lokare et al.}

\begin{abstract}
Feature engineering plays an important role in the success of a machine learning model. Most of the effort in training a model goes into data preparation and choosing the right representation. In this paper, we propose a robust feature engineering method, Randomized Union of Locally Linear Subspaces (RULLS). We generate sparse, non-negative, and rotation invariant features in an unsupervised fashion. RULLS aggregates features from a random union of subspaces by describing each point using globally chosen landmarks. These landmarks serve as anchor points for choosing subspaces. Our method provides a way to select features that are relevant in the neighborhood around these chosen landmarks. Distances from each data point to $k$ closest landmarks are encoded in the feature matrix. The final feature representation is a union of features from all chosen subspaces.

The effectiveness of our algorithm is shown on various real-world datasets for tasks such as clustering and classification of raw data and in the presence of noise. We compare our method with existing feature generation methods. Results show a high performance of our method on both classification and clustering tasks. 
\end{abstract}
%
%



\maketitle
\section{Introduction}
\label{sec:Introduction}

The success of a machine learning model depends heavily on the data representation that is fed into the model. Often, the features in the original data are not optimal and it requires feature engineering to learn a good representation \cite{Domingos:2012:FUT:2347736.2347755}. Recent work in feature selection methods \cite{chandrashekar2014survey,guyon2003introduction} and feature engineering methods \cite{Wang:2017:RFE:3097983.3098001} further highlight the importance of giving the right input to machine learning models. Additionally, recent advancements in domain specific feature engineering methods in areas of text mining \cite{forman2003extensive}, speech recognition \cite{seide2011feature}, and emotion recognition \cite{kim2013deep} have shown promising results.
\begin{figure}[t!]
	\centering
	\includegraphics[width=\columnwidth]{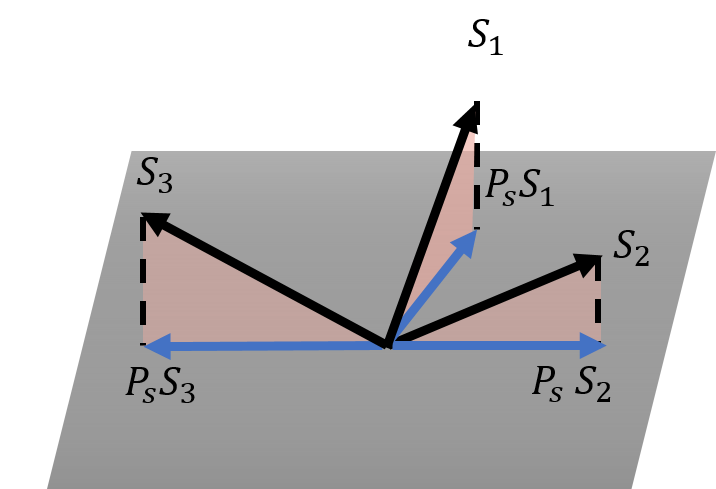} 
	\caption{Example of union of three subspaces and their projections onto a lower dimensional space.}
	\label{fig:Subspace}
\end{figure}
Analyzing high-dimensional datasets can be challenging and computationally expensive. These datasets usually have some features that are irrelevant to the task at hand. Selecting features appropriately reduces the dimensionality and correlation between the features which is seen to improve
classification and clustering performance. Several techniques have been developed to reduce the dimensions (features) of the input data. Dimension reduction techniques are broadly classified as linear and nonlinear approaches. Linear dimension reduction approaches assume data points lie close to a linear (affine) subspace in the input space. Such methods globally transform the data by rotation, translation, and/or scaling. Non-linear methods, sometimes referred to as manifold learning approaches, often assume that input data lies along a low dimensional manifold embedded in a high dimensional space. In this paper, we use a piecewise-linear model defined by a union of subspaces (see Section \ref{sec:UoS}), thus enabling us to handle non-linearly distributed data via locally linear approximations.  

The most commonly used linear dimension reduction method is PCA \cite{jolliffe2002principal}. PCA projects the data onto a linear subspace by finding the directions along which the data has maximum variance as well as the relative importance of these directions. This can be realized through singular value decomposition (SVD). The eigenvectors corresponding to the largest eigenvalues of the covariance matrix are the principal components. More formally, consider $X = [x_1, x_2, \cdots x_N]$. Let the orthogonal basis which maximizes the variance be $y_i = U^T x_i$. It can be shown that $U$ can be obtained from the first $l$ eigenvectors of $\sum = X^T X$, as $Cov(Y) = U^T \sum U$. We use these principal components to project our data. Linearly projecting data to a subspace allows for a mapping between the original space and the new space.
\begin{figure*}[!t]
	\centering
	\includegraphics[width=2.1\columnwidth]{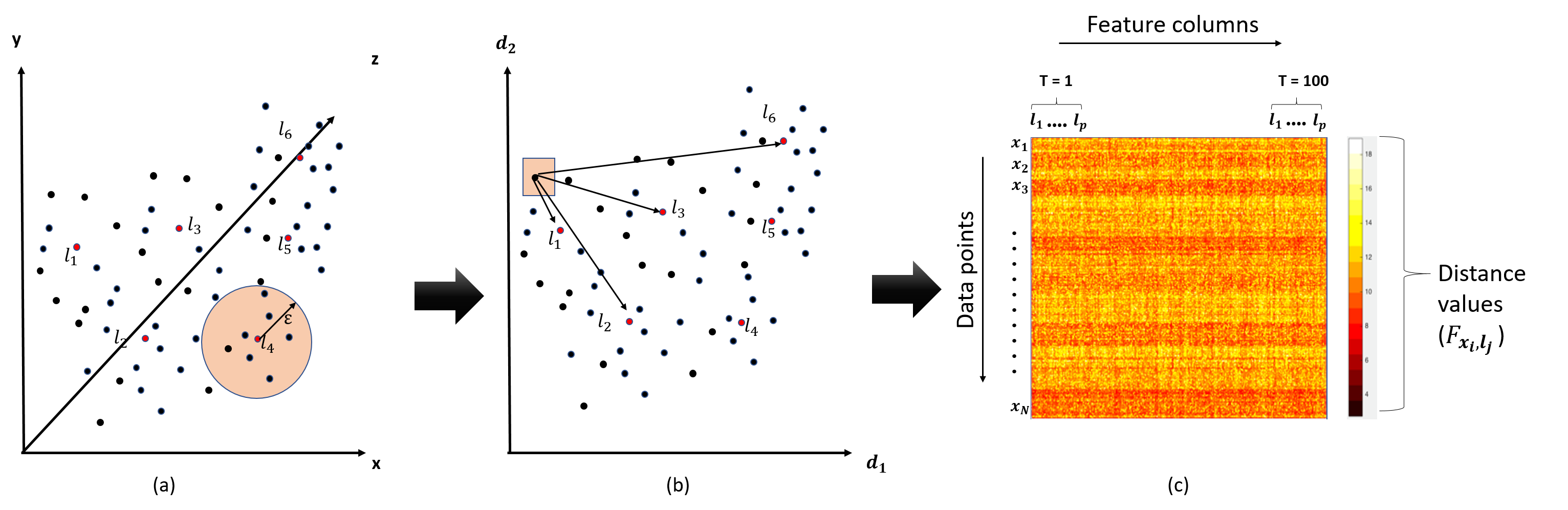} 
	\caption{RULLS for Feature Engineering. (a) Picking landmarks ($l_1,l_2, \cdots, l_6$) in the original space(in this case $\mathbb{R}^3$) with $\epsilon$ neighborhood  (b) Finding closest landmarks after projecting onto the new subspace ($\mathbb{R}^2$ ) (c) Sparse feature matrix after concatenation of features from all the selected subspaces.}
	\label{fig:Pipeline}
\end{figure*}
In our case, we assume that our dataset lives on or close to the union of linear subspaces of low dimension. Our main contribution in this work is to show how we can use a union of subspaces, project data to these subspaces and extract robust and sparse features which, thanks to the properties of random projections in the presence of inherently sparse data, are highly effective. The generated features are not only discriminative but when used in conjunction with simple models are also faster to train. In Subsection \ref{sec:UoS}, we discuss briefly the concept of Union of Subspaces. 

\subsection{Union of Subspaces}
\label{sec:UoS}

The concept of Union of Subspaces (UoS) has been used widely in the context of compressed sensing and sampling \cite{5290295,4483755, 4802322}. It has been shown that many signals of interest can be approximated by a union of subspaces. We use this concept and project the data onto multiple random locally linear subspaces to learn robust features. We consider finite unions of finite dimensional spaces to generate sparse features. Let $X = \{x_{1}, x_{2}, \cdots x_{N}\} \in \mathbb{R}^m$ be the dataset of interest with $N$ data points and $m$ dimensions. These data points can be considered to live in a finite union of subspaces (i.e. there are $\Gamma$ subspaces and $\Gamma < \infty$) which is defined as follows,
\begin{equation}
\chi = \bigcup\limits_{\gamma=1}^{\Gamma} S_{\gamma}
\end{equation}

Figure \ref{fig:Subspace} shows an example of  a union of three subspaces $S_1, S_2,$ and $S_3$. The data are mapped to these subspaces via linear projection. There is an invertible mapping between $X$ and $P_S X$ as long as no two subspaces are projected to the same line on $S$. 

In this work, our focus is not on reconstructing the original data, but in obtaining a sparser representation by identifying locally relevant subspaces. These subspaces are disjoint and low dimensional compared to the dimension of the original space. Once projected to these local subspaces, points in the dataset are described by distances from the nearest landmarks in each subspace. These landmarks are chosen randomly and the neighborhood around them defines the subspace for each landmark. Then we encode distances (these are our features) with respect to these global landmarks. The final feature representation is a union of features from all the chosen subspaces.  

\subsection{Paper Organization}
The remainder of this paper is organized as follows: we present our proposed method in Section \ref{sec:ProposedMethod}; variants of our method are described in \ref{sec:variants}; in Section \ref{sec:Experiments} we show results on real world data sets and compare it with proposed variants and existing state-of-the-art methods. Section \ref{sec:Interpretability} discusses the characteristics of the estimated features, followed by conclusions in Section \ref{sec:Conclusion}.

\section{Proposed Method}
\label{sec:ProposedMethod}
In this section we present our novel method to generate sparse features that can be used for classification or clustering tasks. The proposed method finds a union of subspaces and generates features that locally describe data points with respect to globally chosen landmarks. The motivation behind this idea is that describing data points with local information can provide a robust characterization of the data points. Essentially, the concatenation of local subspaces constitutes an overcomplete dictionary. Most real-world data admits a sparse representation, and compressed sensing theory shows that random projection strategies can be highly effective under such a dictionary. Summarizing this information with respect to globally chosen random landmarks can improve performance in classification and clustering tasks.

The pipeline for our RULLS method is shown in Figure \ref{fig:Pipeline}. RULLS looks at neighborhoods around randomly chosen landmarks (see Figure \ref{fig:Pipeline}(a)). Linear subspace analysis is then performed on the neighborhood chosen around these landmarks. Features that are relevant to this neighborhood are used to project the input dataset to this new subspace (see Figure \ref{fig:Pipeline} (b)). Distances are encoded in a sparse feature matrix comprising of $k$ landmarks for each data point. The resulting features are non-negative and the sparsity of features is controlled by the number of landmarks chosen (see Figure \ref{fig:Pipeline} (c)). By incorporating subspace analysis and choosing appropriate features we obtain a better performance with fewer iterations than alternative methods. We provide details about RULLS in Section \ref{sec:Subspace}. 

\subsection{Randomized Union of Locally Linear Subspaces (RULLS)}
\label{sec:Subspace}
Here we present details of our algorithm. 
\subsubsection{Problem Formulation}
Consider a data matrix $X \in \mathbb{R}^{n \times m}$. Our objective is to create a matrix $F\in \mathbb{R}^{N \times l_p T}$ of features, defined in a manner similar to [26]. This is done iteratively, with iteration index $t=1,\ldots,T$. At each iteration $t$, we choose $l_p$ landmarks (points randomly chosen from the dataset). Around each landmark, we estimate a subspace via SVD. Hence, at iteration $t$ we constitute a UoS as follows:
\begin{equation}
\chi_{t} = \bigcup\limits_{\gamma=1}^{lp} S_{\gamma}
\end{equation}
where $l_p$ is the number of landmarks and $S_{\gamma}$ is the subspace per landmark. The dimension of each subspace depends on the number of features that are sufficient to describe a neighborhood around the landmark.

At iteration $t$, the feature matrix is augmented via concatenating the features from the projection onto each subspace. The number of closest landmarks to a data point chosen to encode distances controls the sparsity and the robustness of the features. We denote this parameter as $l_k$. The sparsity of the feature matrix will be defined as follows,
\begin{equation}
\textit{Sparsity} = \sum_{t =1}^{T} \sum_{i=1}^{N} X_{i} \cdot l_{k}
\end{equation}
where T is the number of iterations, N is the number of data points in a dataset and $X$ is our dataset. 
We also define a Sparsity Ratio (SR) for a given dataset which depends on the number of nearest landmarks ($l_k$), total landmarks ($l_p$), iterations (T), and data points in the dataset (N). The SR is defined as follows, 
\begin{equation}
\textit{Sparsity Ratio (SR)} = \frac{\textit{Sparsity}}{N \cdot l_p \cdot T}
\end{equation}
The sparsity ratio tells us how many elements in the feature matrix are non-zero. SR $\in [0,1]$, where 0 corresponds to empty and 1 to completely filled. The parameter $l_k$ for our approach is chosen as follows,
\begin{equation}
1 < l_k < l_p < N 
\end{equation}  

The goal of the algorithm is to achieve a level of sparsity that will ensure that features are descriptive and robust. In Subsection \ref{sec:alg_rulls} we discuss our algorithm in detail.

\subsubsection{Algorithm}
\label{sec:alg_rulls}
Given an input dataset $X = \{x_1, x_2, \cdots x_n \} \in \mathbb{R}^{m}$, randomly pick $l_p$ landmarks from $X$. We consider a neighborhood around each landmark to get relevant features that describe points in this neighborhood. Once we learn the subspace of this neighborhood we project the dataset to this subspace. We then find Euclidean distances between every point to landmarks in this new space. Next, we find the $l_k$ closest landmarks to every point $x_i$ and fill the feature matrix at those locations as follows,

\begin{equation}
F\{x_i,l_j\} = max ((Mean(D_{x_i}) - D(x_i,l_j)), reg_p \cdot Mean(D_{x_i})) 
\end{equation}

where, $x_i$ corresponds to a data point  $\in X$and $l_j$ corresponds to the $j$ th nearest landmark to $x_i$.  $Mean(D_{x_i})$ is the average distance of all the landmarks to $x_i$ and $D(x_i,l_j)$ is the euclidean distance from $x_i$ to $l_j$. We add a regularization term ($reg_p$) to reduce the effect of outliers. $F\{x_i,l_j\}$ captures the local information with respect to these globally chosen landmarks.
\begin{algorithm}
		\caption{RULLS}
		\label{alg:Subspace Feature Engineering}
		\begin{algorithmic}[1]
			\State {Input: Dataset $X$ $\in$ $\mathbb{R}^{n \times 					   m}$
			      }
			\For{t = 1 to T}
			\State {Choose landmarks $l_p$ randomly from $X$} 
			\For  {each landmark $l_p$}
			\State {Consider an $\epsilon-$ ball around $l_p$}
			\State {Choose $k_{\epsilon}$ neighbors of $l_p$ from 						this neighborhood
				  }
			\State {Find local subspace distances using algorithm 					  \ref{alg:LocalSubspace} and return $D_p$
				  }
			\State {Aggregate $D_p$ to $D$}
			\EndFor
			\State {Use $D$ to find $l_k$ nearest landmarks to 							each data point in $X$
				  }
			\For {each data point $x_i$ in $X$}
			\For {each nearest landmark $j \in (l_1, \cdots, l_k)$}
			\State {Set $F\{x_i,l_j\} = max(Mean(D_{x_i}) - D(x_i,l_j), \ 							reg_{p} \cdot Mean(D_{x_i}))$
				  }
			\EndFor
			\EndFor
			\EndFor
			\State {Return concatenated features $F$}
		\end{algorithmic}
\end{algorithm}

The value of T is experimentally decided in our algorithm and based on the experiments we find that T in the range of 1 to 100 is sufficient to get a good performance. The best choice on the number of landmarks $l_p$ depends on the dataset. Since we do not know the appropriate amount of sparsity for a dataset and we assume that datasets have at least 1024 data points, number of landmarks $l_p$ are selected based on $min(2^{log_{2}(N/2)}$ , $2^{log_{2}(1024)})$, as specified in \cite{Wang:2017:RFE:3097983.3098001}, where N is the number of data points in the dataset. 
\begin{algorithm}
		\caption{Local Subspace Distances}
		\label{alg:LocalSubspace}
		\begin{algorithmic}[1]
			\State Input: Dataset $X$ $\in$ $\mathbb{R}^{n \times 					   m}$,  Neighborhood $X_{\epsilon}$ $\in$ $\mathbb{R}^{k_{\epsilon} \times m}$,  and landmark $l_p$
			\If {flag = 1}
			 	\State {Normalize $X_{\epsilon}$ with respect to the neighborhood}
			 	\State {Normalize $X$ with respect to the neighborhood's subspace}
			\EndIf 	 	
			\State Compute the eigenvalues and eigenvector of the covariance matrix of this neighborhood
			\State Compute the dimensions that explain $95 \%$ of the variance of the data
			\State Project $X$ to this subspace and compute distances to the landmark $l_p$ to create $D_p$ $\in$ $\mathbb{R}^{n \times l_p}$
			\State Return $D_p$
		\end{algorithmic}
\end{algorithm}

The dimensions of the data points vary based on the chosen subspace. This allows us to throw away irrelevant features and choose features in a better way. For the number of neighbors $k_\epsilon$, based on the experiments a value $k_\epsilon$ can be in the range of $10 - 30$. These points determine the subspace to project the dataset onto and the number of features. The number of nearest landmarks $l_k$ is generally chosen $1< l_k < l_p < N$, to improve robustness and reduce the effect of outlier landmarks. 

Normalization of the data is optional. The normalization step is skipped if the dataset was normalized during pre-processing or if normalization is not required (set flag to 0). If flag is 1, then normalization is performed on the input data with respect to the neighborhood. We use a regularization parameter $reg_{p} = 0.0001$ to smooth and control effect of outlier landmarks. 

The resulting feature matrix $F$ is the aggregation of the features at every iteration $t$. At every iteration a different UoS is chosen and features are concatenated. $F$ is sparse and has non-zero entries only at landmark points close to a data point.  

\subsection{Addressing Outliers while choosing Subspaces}
Principal component analysis is one of the most widely used technique for dimension reduction. It has been widely used in applications in signal processing, image processing, and pattern recognition. It is well known that PCA is sensitive to outliers and noise. In order to overcome this shortcoming various incremental alternatives have been suggested in the literature \cite{xu1995robust, 937541,verboon1994resistant}. However, these iterative methods are computationally expensive due to an iterative solution to the optimization problem. 
 
Robust approaches based on projection pursuit have been introduced which can handle high-dimensional data. ROBPCA was introduced by Hubert \etal \cite{doi:10.1198/004017004000000563} which combines projection pursuit ideas with robust scatter matrix estimation. We use this method in our work due to faster computation and robustness to outliers.

In our experiments, we use regular PCA and also show examples where we replace the PCA with the ROBPCA \cite{verboven2010matlab, doi:10.1198/004017004000000563}. Depending on the type of noise added we see a difference in performance between using PCA and ROBPCA. 

\subsection{Time and Space Complexity}
\textbf{Time Complexity}: Consider a dataset $X \in R^{n \times m}$.
For $t$ th iteration, we find local subspaces over a small neighborhood with points $k_\epsilon$ constant. The cost of computing the covariance matrix is $O(k_\epsilon \  m^2)$ which is $O(m^2)$ due to small $k_\epsilon$. The cost of performing the eigenvalue decomposition and selecting top ranked basis is $O(m^3)$. So the total cost of selecting the local subspace is $O(l_p^{t}\ m^3)$. This cost is dependent on the dimension of the dataset.  

The cost of computing distance from $n$ data points to landmarks depends on the number of features chosen per subspace $m_p^{t}$ and the number of landmarks $l_p^{t}$, which is given by $O(n\  m_p^{t} \ l_p^{t})$. The number of features chosen after selecting a local subspace ($m_p^{t}$) will always be less than equal to the number of features in the raw features ($m$) ($m_p^{t} \leq m $) due to our choice of threshold to select top ranked basis. The cost of computing the mean for each data point $i$ in dataset $X$ is $O(n \ l_p^{t})$.

The total cost of the algorithm is $O(\sum_{t=1}^{T} l_p^{t} \  ( m^3 + n \  m_p^{t}))$. Selecting the locally relevant features is the most expensive part in the algorithm. This cost can be reduced to $(m^2)$ by using modifications of the QR based approaches \cite{2016arXiv161110142A}. \\

\textbf{Space Complexity}: For a input dataset $X \in R^{n \times m^t}$, the computation of distance matrix $D \in R^{n \times l_{p}^{t}} $ takes $O( n \  l_{p}^{t})$ space. The aggregated features $F_t \in R^{n \times l_{p}^{t}}$  for iteration $t$ constitutes a sparse matrix. The space complexity for $F_t$ is $O(n \ l_{p}^{t})$. Therefore, the total space complexity for RULLS is $O(\sum_{t=1}^{T} n \ l_{p}^{t})$ which is $O(n \ l_{p} \ T )$.

\section{Variations of RULLS}
\label{sec:variants}
In this section we present other ways to generate features. Random projections are used in the literature frequently in applications such as dimension reduction \cite{Bingham:2001:RPD:502512.502546} and compressed sensing \cite{candes2006near}. In our case we use random projections to project the data onto random subspaces. The number of features selected per iteration is kept constant in this approach. The final feature is the aggregation of features from the union of these random subspaces. We call this method Variant \Romannum{1}.

The Johnson-Lindenstrauss Lemma \cite{johnson1984extensions,dasgupta1999elementary} states that given a set $S$ of points in $R^n$, if we perform an orthogonal projection of those points onto a random $d$-dimensional subspace, then $d = O( \frac{1}{\gamma^2} log |S|)$ is sufficient so that with
high probability all pairwise distances are preserved up to $1 \pm \gamma$ (up to scaling).

\begin{algorithm}
		\caption{Variant \Romannum{1}}
		\label{alg:Random projections}
		\begin{algorithmic}[1]
			\State Input: Dataset $X$ $\in$ $\mathbb{R}^{n \times m}$
			\For{t = 1 to T}
			    \State {Randomly project the data $X$ to $X_t$ using $d$ dimensions}
			    \State {Compute features $F_t$ using the algorithm \ref{alg:Distance_proj}} 
			    \State {Aggregate features $F_t$ to $F$.}
			\EndFor
			\State{return F}
		\end{algorithmic}
\end{algorithm}

We use this property of random projections and propose a modification to RULLS to generate sparse features. In this approach we fix the number of dimensions $d$ (this is constant for all subspaces). We use a mapping $R^m \rightarrow R^d$ and project the data to $d$ dimensions. All further computations to find landmarks are done in the projected space. The number of dimensions in the projected space $d$ varies between $10 \%$ and $40\%$ of $m$, where $m$ is the dimensionality of the original dataset. Variant \Romannum{1} is described in Algorithms \ref{alg:Random projections} and \ref{alg:Distance_proj}. The feature values are computed in a similar manner as in the RULLS method.

\begin{algorithm}
		\caption{Compute features}
		\label{alg:Distance_proj}
		\begin{algorithmic}[1]
			\State Input: $X_t$ $\in$ $\mathbb{R}^{n \times d}$ 
			\State Randomly select $p$ landmarks $l_1, \cdots, l_p$
            \For{each data point $x_i$ in $X_t$}
            \State Calculate the distance to each landmark
			\State Find $k$ nearest landmarks $l_{1},\cdots, l_{k}$
			\For{each neighbor $j \in (l_{1},\cdots, l_k )$}
			\State Set $F\{x_i,l_j\} = max(Mean(D_{x_i}) - D(x_i,l_j), reg_{p} \cdot Mean(D_{x_i}))$
			\EndFor
			\EndFor
			\State Return $F$
		\end{algorithmic}
\end{algorithm}

Prior work in randomized feature engineering (RandLocal) was introduced in \cite{Wang:2017:RFE:3097983.3098001}. We implement the following modifications to the RandLocal method and call this method Variant \Romannum{2}\\
\begin{itemize}
\item{We increase the number of nearest neighbors to make the method robust to outliers.}
\item{RandLocal finds only one nearest landmark, i.e. $k = 1$. If $l_1$ happens to coincide with the data point $x_i$, then $l_1$ is replaced by the next nearest landmark. We found that this assignment resulted in misclassification rate to go higher, so in our variant we avoid this by considering more landmarks instead of just one.}
\end{itemize}

The basic idea is that a point can be better described by global points of interest instead of just using one local descriptor. Variant \Romannum{2} is described in Algorithms \ref{alg:Variant of RandLocal} and \ref{alg:Distance}.
\begin{algorithm}
		\caption{Variant \Romannum{2}}
		\label{alg:Variant of RandLocal}
		\begin{algorithmic}[1]
			\State Input: Dataset $X$ $\in$ $\mathbb{R}^{N \times d}$
			\For{t = 1 to T}
			    \State {Create $X_t$ $\in$ $R^{N \times m}$  by randomly sampling over the dimensions of $X$}
			    \State {Compute features $F_t$ using the algorithm \ref{alg:Distance}} 
			    \State {Aggregate features $F_t$ to $F$.}
			\EndFor
			\State{return F}
		\end{algorithmic}
\end{algorithm}

\begin{algorithm}
		\caption{Compute features}
		\label{alg:Distance}
		\begin{algorithmic}[1]
			\State Input: $X_t$ $\in$ $\mathbb{R}^{N \times m}$ 
			\State Randomly select $p$ landmarks $l_1, \cdots, l_p$
            \For{each data point $x_i$ in $X_t$}
            \State Calculate the distance to each landmark
			\State Find $k$ nearest landmarks $l_{1}, \cdots, l_{k}$
			\For{each neighbor $j \in (l_1, \cdots, l_k ) $ }
			\State Set $F\{x_i,l_j\} = max(Mean(D_{x_i}) - D(x_i,l_j), reg_{p} \cdot Mean(D_{x_i}))$
			\EndFor
			\EndFor
			\State Return $F$
		\end{algorithmic}
\end{algorithm}

The drawback of Variant \Romannum{1} (\ref{alg:Random projections}, \ref{alg:Distance_proj}), Variant \Romannum{2} (\ref{alg:Variant of RandLocal}, \ref{alg:Distance}) and RandLocal \cite{Wang:2017:RFE:3097983.3098001} is that these methods do not consider local information to choose the number of dimensions. They randomly pick features which is not the most effective way to select relevant features. Furthermore, they may require additional parameter constraints to retain the relevant features.

\section{Experiments}
\label{sec:Experiments}
In this section, we describe the datasets used and our experimental setup for each of the methods compared. 
\begin{table*}[h]
\centering
\caption{Description of datasets used for classification and clustering experiments}
\label{dataset}
\resizebox{1.5\columnwidth}{!}{\begin{tabular}{|cccccccc|}
\hline
Dataset & \begin{tabular}[c]{@{}c@{}}Japanese\\ Vowel\end{tabular} & \begin{tabular}[c]{@{}c@{}}Fashion\\  MNIST\end{tabular} & Baseball & \begin{tabular}[c]{@{}c@{}}Breast \\ Cancer\end{tabular} & Digits & IRIS & \begin{tabular}[c]{@{}c@{}}Anuran\\ Calls\end{tabular} \\ \hline
Instances & 9,960 & 70,000 & 1,340 & 569 & 10,992 & 150 & 7195 \\ \hline
\#Features & 15 & 784 & 18 & 32 & 16 & 4 & 22 \\ \hline
\multicolumn{1}{|l}{\#Classes} & 9 & 10 & 3 & 2 & 10 & 3 & 4 \\ \hline
Missing & - & - & 20 & - & - & - & - \\ \hline
\end{tabular}}
\end{table*}

\subsection{Datasets}
We used benchmark datasets from OpenML \cite{OpenML2013} and UCI repository \cite{Lichman:2013}. The datasets used are as follows: Fashion-MNIST \cite{xiao2017}, Japanese Vowel \cite{Lichman:2013}, Baseball \cite{simonoff2013analyzing}, Breast Cancer Wisconsin dataset \cite{Lichman:2013}, Anuran Calls \cite{Lichman:2013}, and Digits \cite{Lichman:2013}. The statistics of these datasets are shown in Table \ref{dataset}. These datasets include images, multivariate data, multiple classes and missing values. These datasets are highly imbalanced and are good candidates for our analysis. We show the performance of our generated features for classification, robustness to noise and clustering tasks.
\begin{figure*}[t!]
	\centering
	\includegraphics[width=2\columnwidth]{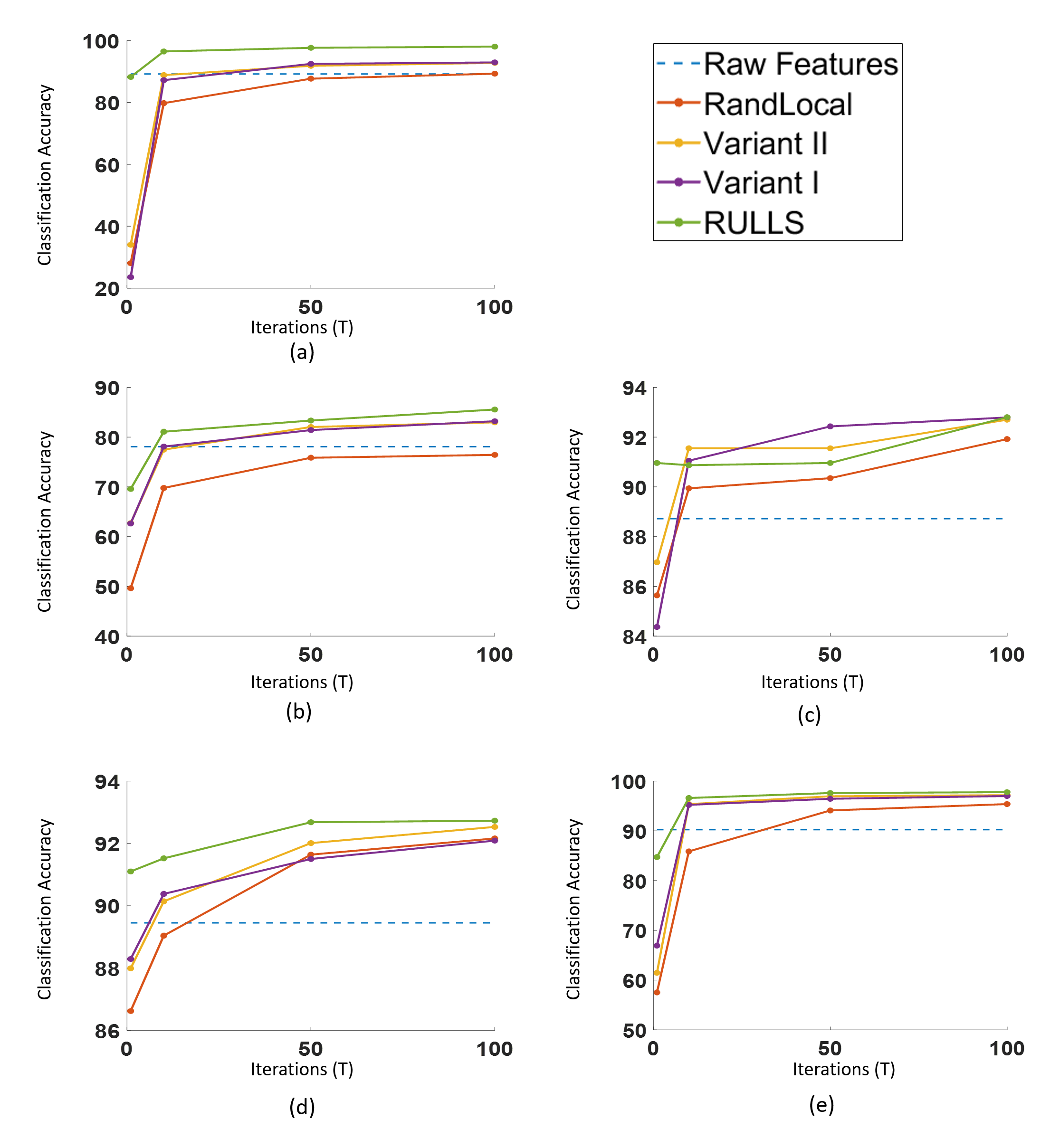} 
	\caption{Average classification accuracy ($\%$) with varying iterations (t = 1, 10, 50, and 100) for raw features, RandLocal, Variant \Romannum{1}, Variant \Romannum{2} and RULLS (PCA). (a) Japanese Vowel, (b) Fashion MNIST, (c) Breast Cancer Wisconsin, (d) Baseball and (d) Digits dataset. Methods compared here beat the raw features score in just a few iterations. RULLS performs better than other methods on all datasets.}
	\label{fig:Iter}
\end{figure*}

\subsection{Comparison}
In this section, we compare raw features, RandLocal, Variant \Romannum{1}, Variant \Romannum{2} and RULLS. The code for our method can be found at \cite{Codesite}. Our datasets were randomly divided into training ($80\%$) and test ($20 \%$). We perform 10-fold cross validation and report the average results. An example of the features generated is shown in Figure \ref{fig:Pipeline}(c). We render the feature matrix generated by using RULLS on the Japanese Vowel dataset for $T = 100$, as an image. The brighter regions correspond to the non-zero regions with distance values in the matrix while the darker regions show empty spaces. For clustering and classification tasks we report results at $T = 100$ for all methods. We notice patterns in the feature matrix which is further analyzed in section \ref{sec:Interpretability}.  
\begin{table*}[!h]
\centering
\caption{Classification accuracy on datasets. Highlighted text shows the method with the best performance.}
\label{classification}
\resizebox{1.2\columnwidth}{!}{\begin{tabular}{|cccccc|}
\hline
Method & \begin{tabular}[c]{@{}c@{}}Japanese\\ Vowel\end{tabular} & \begin{tabular}[c]{@{}c@{}}Fashion\\ MNIST\end{tabular} & \begin{tabular}[c]{@{}c@{}}Breast \\ Cancer\end{tabular} & Baseball & Digits \\ \hline
Raw Features & 89.18 & 78.06 & 88.72 & 89.45 & 90.25 \\ \hline
RandLocal & 89.29 & 76.43 & 91.92 & 92.16 & 95.38 \\ \hline
Variant \Romannum{1} & 92.92 & 83.19 & 92.79 & 92.09 & 97.00 \\ \hline
Variant \Romannum{2} & 92.76 & 82.96 & {\color[HTML]{333333} 92.70} & 92.53 & 97.17 \\ \hline
RULLS (PCA) & {\color[HTML]{3531FF} 98.02} & {\color[HTML]{3531FF} 85.54} & {\color[HTML]{3531FF} 92.80} & {\color[HTML]{3531FF} 92.73} & {\color[HTML]{3531FF} 97.66} \\ \hline
\end{tabular}}
\end{table*}
We set the $\epsilon$ ball radius such that we get $k_\epsilon = 30$. The number of nearest landmarks $l_k$ is set to $10$, this parameter is used in the Variant \Romannum{1}, Variant \Romannum{2} and RULLS. The Randlocal method uses only one nearest landmark so $l_k = 1$. The regularization parameter $reg_{p}$ is set to a small number $0.0001$. The number of features $d$ for RandLocal, Variant \Romannum{1}, and Variant \Romannum{2} method were set to $0.2 \cdot m$, where $m$ is the dimension of each dataset. In case of RULLS, this parameter is chosen by performing the local subspace analysis and it varies per landmark ($d < m$, where $m$ is the dimension of the original dataset and $d$ is the dimension of the reduced space).

We would like to point out some key aspects of RULLS. We consider the importance of features unlike RandLocal, Variant \Romannum{1}, and Variant \Romannum{2} methods where features are picked randomly. This allows us to get a better performance with similar parameter settings and fewer iterations. For all the methods compared with RULLS choosing random dimensions may affect the performance especially in the unsupervised settings. These algorithms may need to be modified to take these scenarios into account. 

\subsection{Classification Performance}
Classification task was performed using a linear Support Vector Machine (SVM) \cite{cortes1995support} classifier. Figure \ref{fig:Iter} shows the performance of the compared methods on different datasets at different iterations. We choose $T = 1$, $10$, $50$, and $100$ iterations to analyze the effect of adding sparse features. RULLS uses regular PCA in this experiment.

\begin{table*}[!h]
\centering
\caption{Classification performance in presence of $10 \%$ noise added to columns and rows in each dataset. Best performance is highlighted in blue. The numbers in the parenthesis indicate the difference between the performance with and without noise.}
\label{noise}
\resizebox{\textwidth}{!}{
\begin{tabular}{|c|llll|llll|}
\hline
 & \multicolumn{4}{c|}{\begin{tabular}[c]{@{}c@{}}Add noise to columns\\ (10\%)\end{tabular}} & \multicolumn{4}{c|}{\begin{tabular}[c]{@{}c@{}}Add noise to rows\\ (10\%)\end{tabular}} \\ \cline{2-9} 
\multirow{-2}{*}{Method} & \multicolumn{1}{c}{\begin{tabular}[c]{@{}c@{}}Japanese\\ Vowel\end{tabular}} & \multicolumn{1}{c}{\begin{tabular}[c]{@{}c@{}}Fashion\\ MNIST\end{tabular}} & \multicolumn{1}{c}{\begin{tabular}[c]{@{}c@{}}Breast \\ Cancer\end{tabular}} & \multicolumn{1}{c|}{Baseball} & \multicolumn{1}{c}{\begin{tabular}[c]{@{}c@{}}Japanese\\ Vowel\end{tabular}} & \multicolumn{1}{c}{\begin{tabular}[c]{@{}c@{}}Fashion\\ MNIST\end{tabular}} & \multicolumn{1}{c}{\begin{tabular}[c]{@{}c@{}}Breast\\ Cancer\end{tabular}} & \multicolumn{1}{c|}{Baseball} \\ \hline
Raw Features & 87.90 (1.28) & 77.79 (0.27) & 80.57 (8.15) & 90.30 (0.85) & 67.53 (21.65) & 79.06 (1.00) & 84.36 (4.36) & 88.81 (0.64) \\ \hline
RandLocal & 84.91 (4.38) & 74.65 (1.78) & 86.78 (5.14) & 91.76 (0.40) & 81.87 (7.42) & 76.46 (0.03) & 90.70 (1.22) & 91.41 (0.75) \\ \hline
Variant \Romannum{1} & 91.16 (1.76) & 82.04 (1.15) & 88.96 (3.83) & 91.34 (0.75) & 85.04 (7.88) & 82.74 (0.45) & 91.39 (1.40) & 91.56 (0.53) \\ \hline
Variant \Romannum{2} & 91.28 (1.48) & 81.85 (1.11) & {\color[HTML]{3531FF} 89.64 (3.06)} & 91.27 (1.26) & 84.80 (7.96) & 82.82 (0.14) & {\color[HTML]{3531FF} 92.45 (0.25)} & 91.33 (1.20) \\ \hline
RULLS (PCA) & {\color[HTML]{3531FF} 91.76 (6.26)} & {\color[HTML]{3531FF} 84.06 (1.48)} & {\color[HTML]{333333} 88.07 (4.73)} & {\color[HTML]{3531FF} 91.79 (0.94)} & {\color[HTML]{3531FF} 89.61 (8.41)} & {\color[HTML]{3531FF} 85.55 (0.01)} & 90.17 (2.63) & {\color[HTML]{3531FF} 91.86 (0.87)} \\ \hline
\end{tabular}
}
\end{table*}

The dotted blue line in Figure \ref{fig:Iter} corresponds to the  classification accuracy of raw features. From Figure \ref{fig:Iter}, we observe that the difference in classification accuracy between $T = 50$ and $T = 100$ for all methods is very small. This indicates that adding more features beyond this point will only lead to very small improvements in the performance. We observe that $T = 100$ gives the best performance for all methods hence we report the results at $T =100$ (see Table \ref{classification}).

In Table \ref{classification}, we note a significant improvement in case of RULLS over the raw features specifically, $8 \%$ for Japanese Vowel dataset, $7.84 \%$ for the Fashion MNIST dataset, $ 4.08 \%$ for Breast Cancer Wisconsin dataset, $3.28 \%$ for Baseball dataset, and $7.41 \%$ for the Digits dataset. 

For all the datsets RULLS performs better than the compared methods. RULLS beats the existing RandLocal by $8.73 \%$ for Japanese Vowel dataset, $9.11 \%$ for the Fashion MNIST dataset, $ 0.88 \%$ for Breast Cancer Wisconsin dataset, $0.12 \%$ for Baseball dataset, and $2.28 \%$ for the Digits dataset. Variant \Romannum{1} and Variant \Romannum{2} perform better than RandLocal for all datasets as well.

\subsection{Robustness to Noise}
In this section we describe the experimental setup to examine the robustness of our proposed method in the presence of noise. We added noise in two ways: corrupting the $10 \%$ columns in the data and corrupting $10 \%$ rows in the data. We added uniform random noise in both cases. Table \ref{noise} shows the classification performance for both cases. The numbers in the parenthesis indicate the difference between the performance with and without noise.  

We see a drop in the performance for all methods in presence of noise. We observe consistent results for all datasets except for the breast cancer dataset. For this dataset we see that Variant \Romannum{2} performs the best in presence of both types of noise. In this case we modified RULLS to use ROBPCA instead of regular PCA. 

The results are reported in Table \ref{robpca}. RULLS with ROBPCA on raw features showed a slightly lower performance than RULLS with PCA indicating that the raw features does not have outliers.

RULLS with ROBPCA shows improved performance when the rows of the dataset are corrupted by noise. This is expected because adding noise to rows simulates the effect of having outliers. We see a $3.16 \%$ improvement over the RULLS without ROBPCA, which even beats RULLS with ROBPCA on the raw features. This indicates that RULLS with ROBPCA is able to deal with outliers (noise) better that just using PCA. RULLS with robust PCA performs better than Variant \Romannum{2} in case of noise added to the rows.

For the case when we add noise to columns, we do not see an improvement when we use RULLS with ROBPCA. This is expected since ROBPCA works well by reducing the effect of the outliers. In the column case we end up changing the description of data points which is different from the adding noise (outliers) in the row case.   



\begin{table}[]
\centering
\caption{RULLS with ROBPCA on the Breast Cancer dataset in the case of raw features and $10 \%$ noise added to columns and rows.}
\label{robpca}
\resizebox{0.9\columnwidth}{!}{\begin{tabular}{|clcc|}
\hline
\begin{tabular}[c]{@{}c@{}}Breast Cancer \end{tabular} & Raw features & \begin{tabular}[c]{@{}c@{}}Columns\\ (10\%)\end{tabular} & \begin{tabular}[c]{@{}c@{}}Rows\\ (10\%)\end{tabular} \\ \hline
\begin{tabular}[c]{@{}c@{}}RULLS\\ (ROBPCA)\end{tabular} & \multicolumn{1}{c}{92.28} & 86.49 & 93.33 \\ \hline
\end{tabular}}
\end{table}

\subsection{Clustering Performance}
For clustering we use $k$-means algorithm \cite{hartigan1979algorithm}. We report the average Normalized Mutual Information (NMI) for each dataset. For this experiment we use RULLS with regular PCA. We observe that the RULLS performs best for Anuran Calls and Baseball datasets. For the Iris dataset, Variant \Romannum{2} performs the best. 

We tried the ROBPCA in the case of the IRIS dataset. Table \ref{robpca_iris} shows improved performance over regular PCA. We note that the use of ROBPCA is dependent on the type of dataset at hand. In the case of the IRIS dataset we see an improvement of $2.59 \%$ which is comparable to Variant \Romannum{1} and Variant \Romannum{2}.
 
\begin{table}[!t]
\centering
\caption{Clustering performance on datasets. We report Normalized Mutual Information (NMI). Highlighted text shows the method with the best performance per dataset.}
\label{clustering}
\resizebox{0.9\columnwidth}{!}{\begin{tabular}{|cccc|}
\hline
Method & Anuran Calls & IRIS & Baseball \\ \hline
Raw Features & 0.4215 & 0.7582 & 0.1638 \\ \hline
RandLocal & 0.4028 & 0.6523 & 0.1532 \\ \hline
Variant \Romannum{1} & 0.4333 & 0.7980 & 0.1745 \\ \hline
Variant \Romannum{2} & 0.4413 & {\color[HTML]{3531FF} 0.8057} & {\color[HTML]{333333} 0.1907} \\ \hline
RULLS (PCA) & {\color[HTML]{3531FF} 0.4472} & {\color[HTML]{333333} 0.7612} & {\color[HTML]{3531FF} 0.1924} \\ \hline
\end{tabular}}
\end{table}

\begin{table}[!t]
\centering
\caption{Comparison of RULLS with PCA and ROBPCA on IRIS dataset. We see an improvement in performance with ROBPCA}
\label{robpca_iris}
\resizebox{0.9\columnwidth}{!}{\begin{tabular}{|clc|}
\hline
IRIS dataset & RULLS (PCA) & \begin{tabular}[c]{@{}c@{}}RULLS (ROBPCA)\end{tabular} \\ \hline
NMI & \multicolumn{1}{c}{0.7612} & 0.7981 \\ \hline
\end{tabular}}
\end{table}

\section{Feature Analysis}
\label{sec:Interpretability}
In this Section, we attempt to understand the features by visually inspecting them to look for patterns or to find some intuition of how these features compare with each other in terms of their predictive power. 
\begin{figure*}[t!]
	\centering
	\includegraphics[width=2.1\columnwidth]{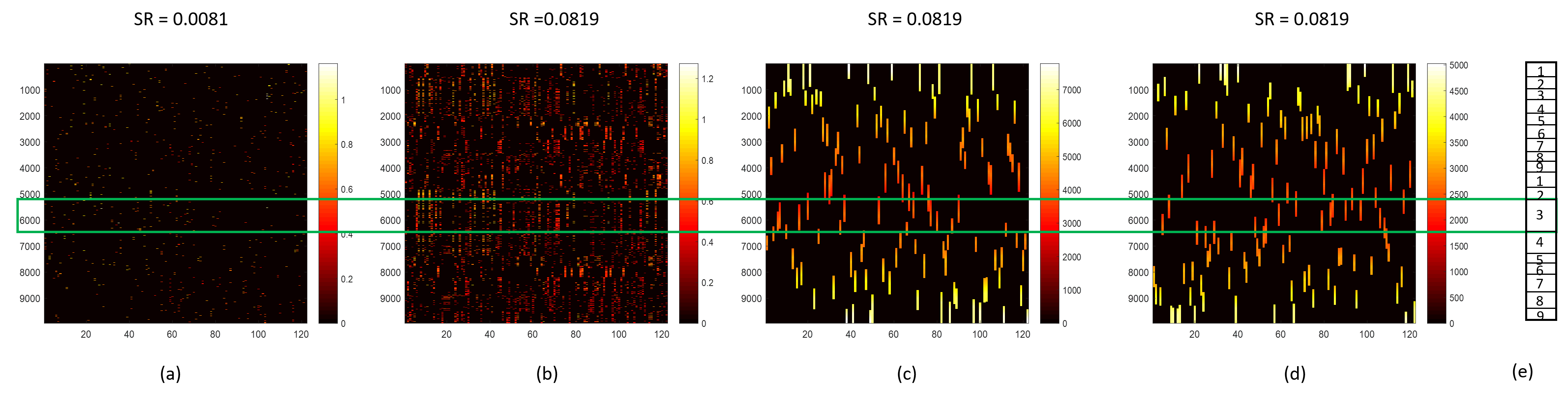} 
	\caption{Visual Interpretation of Features. Rendered images of features for (a) RandLocal, (b) Variant \Romannum{2}, (c) Variant \Romannum{1}, (d) RULLS (PCA), and (e) shows the ground truth class labels. Highlighted green segment shows an example for data points belonging to a class (label = 3) in the Japanese Vowel dataset in all the feature images. Sparsity Ratio (SR) is indicated at the top of each image. Noisy patterns are observed for RandLocal and Variant \Romannum{2} while refined patterns are observed for Variant \Romannum{1} and RULLS (PCA). The features are generated for $T = 1$, $l_p= 122$, $l_k = 10$ for RULLS (PCA), Variant \Romannum{1}, and Variant \Romannum{2}, and $l_k = 1$ for RandLocal.}
	\label{fig:interpret}
\end{figure*}

Figure \ref{fig:interpret} shows the features from (a) RandLocal, (b)  Variant \Romannum{2}, (c)  Variant \Romannum{1}, and (d) RULLS (PCA) for the Japanese Vowel dataset. The parameters for this example are $T = 1, l_p = 122$, $l_k = 10$ for RULLS, Variant \Romannum{1}, and Variant \Romannum{2} and $l_k = 1$ for RandLocal. Figure \ref{fig:interpret} (e), shows each data point with the corresponding classes (in this dataset we have 9 classes). We choose a segment (highlighted in green) as an example to illustrate our observations. This segment belongs to a single class (class label = 3). We note that in Figure \ref{fig:interpret} (a), the feature matrix is very sparse (SR = 0.0081). We observe that points belonging to the same class do not have same neighbors. We suspect that this is due to assigning each data point to only one nearest landmark. In Figure \ref{fig:interpret} (b), the effect of assigning a data point to multiple landmarks can be seen. We observe the feature matrix is less sparse (SR = 0.0819) than in Figure \ref{fig:interpret}(a), however the image appears noisy. Note the patterns in the matrix for points belonging to the same class (see highlighted green segment in Figure \ref{fig:interpret}(b)).

In Variant \Romannum{1} and RULLS (PCA) (Figure \ref{fig:interpret} (c) and (d)), we see refined patterns that are less noisy. Particularly in the highlighted segments we see the data points belonging to same class show solid vertical lines indicating that they pick the same landmarks (neighbors). Notice the range of distances in Figure \ref{fig:interpret} (c) and (d) are in the projected space. 
Similar patterns are seen for points belonging to the same class in these two images which suggests good predictive power. Results from Table \ref{classification} and \ref{clustering} further validate these observations.

The Sparsity Ratio (SR) for each method is shown in the Figure \ref{fig:interpret}. RandLocal has the lowest sparsity ratio while RULLS (PCA), Variant \Romannum{1}, and Variant \Romannum{2} have the same sparsity ratio. Our observations based on the sparsity ratio suggest that there is a trade-off between predictive power and sparsity.


\section{Conclusion}
\label{sec:Conclusion}
The success of machine learning models depends heavily on the features that we feed to it. In this paper we present our unsupervised method (RULLS) to generate robust features. These features are sparse and fast to compute. The raw features are projected to local subspaces by choosing the most descriptive variables in the local neighborhoods. This has an added advantage over choosing features randomly. We also show that by choosing the features using local neighborhoods we can achieve a better performance with fewer iterations. 

We provide modifications to RULLS and further compare all with an existing method RandLocal. RULLS and its variants perform superior to the existing RandLocal method. Experiments indicate that our methods require fewer iterations to give a better performance than the raw features. We suggest using a robust PCA for datasets that have outliers, missing values or noisy samples.

By visually inspecting the extracted features, we gain a better understanding of the differences between the methods. We observe clear patterns for RULLS, which provides intuition for its superior performance.

In this work we consider Euclidean distance but other distances can be used. In future, we would like to test our approach with other linear as well as non-linear dimension reduction methods. We also plan to characterize the relationship between sparsity ratio and predictive power.

\bibliographystyle{ACM-Reference-Format}

\end{document}